Preprint, published in *2022 IEEE International Conference on Big Data (Big Data)*, Osaka, Japan, 2022, pp. 3735-3744, https://doi.org/10.1109/BigData55660.2022.10020507. © 2023 by IEEE.# Discovering Limitations of Image Quality Assessments with Noised Deep Learning Image Sets

Wei Dai
Department of Computer Science
Purdue University Northwest
Hammond, Indiana, USA
weidai@pnw.edu

Daniel Berleant
Department of Information Science
University of Arkansas at Little Rock
Little Rock, Arkansas, USA
jdberleant@ualr.edu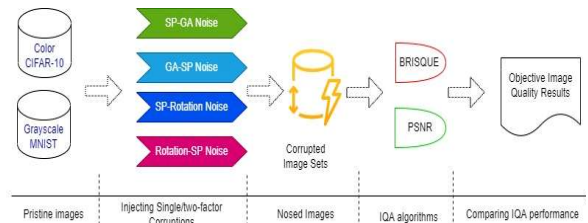

Figure 1. Data Flow

*Abstract*— Image quality is important, and can affect overall performance in image processing and computer vision as well as for numerous other reasons. Image quality assessment (IQA) is consequently a vital task in different applications from aerial photography interpretation to object detection to medical image analysis. In previous research, the BRISQUE algorithm and the PSNR algorithm were evaluated with high resolution (≥ 512×384 pixels), but relatively small image sets (≤4,744 images). However, scientists have not evaluated IQA algorithms on low resolution (≤32×32 pixels), multi-perturbation, big image sets (for example, ≥60,000 different images not counting their perturbations). This study explores these two IQA algorithms through experimental investigation. We first chose two deep learning image sets, CIFAR-10 and MNIST. Then, we added 68 perturbations that add noise to the images in specific sequences and noise intensities. In addition, we tracked the performance outputs of the two IQA algorithms with singly and multiply noised images. After quantitatively analyzing experimental results, we report the limitations of the two IQAs with these noised CIFAR-10 and MNIST image sets. We also explain three potential root causes for performance degradation. These findings point out weaknesses of the two IQA algorithms. The research results provide guidance to scientists and engineers developing accurate, robust IQA algorithms. All source codes, related image sets, and figures are shared on the website (https://github.com/caperock/imagequality) to support future scientific and industrial projects.

*Keywords—Image quality assessments, BRISQUE, PSNR, Quality assessments*## I. INTRODUCTION

In the big data era, digital images are widely used [1]–[4]. Drones and satellites take aerial images daily. Surveillance camera systems enhance public safety. X-ray and computed tomography (CT) images of patients support medical applications. Machine vision and deep learning algorithms use images such as those to train neural networks, detect real objects, and infer rational actions [5]. However, we cannot always capture high-quality, high-resolution images in the real world. For example, drones usually capture blurred images. Yet low quality images impact artificial intelligence applications. For example low quality images can reduce the accuracy of deep neural networks [6], [7]. Therefore, image quality assessment (IQA) algorithms are vital to real world applications.

Existing IQA algorithms are usually evaluated with high resolution images [8]. For example, images of TID2008/2013 are 512 × 384 pixels in size [9], [10]. Much less is known about the performance of IQA algorithms applied to low resolution images with multiple perturbations. In the real world, two-factor corrupted images are often encountered: faded color traffic signals installed on leaning poles may mislead unmanned vehicles because the resulting degraded, rotated pictures reduce the accuracy of deep learning networks.

Previous research explored the performance of IQA algorithms on small image sets corrupted by multi-perturbation conditions [8]. For instance, the LIVEMD database includes merely 15 reference images [11], and MDID2013 set has 12 original images in total [12].

Major image quality assessment databases include 40 different data sets [8], yet rarely are these used for deep learning research projects. For example, researchers have not evaluated IQA algorithms on big datasets with color images and monochrome deep learning datasets: CIFAR-10 has 110 thousand color images; MNIST has 70,000 grayscale images. Obviously, there is a gap between IQA researchers and deep learning scientists even though they are both doing research on image sets.

In this present study, our contributions are as follows.

1) To benchmark the robustness of IQA, we provide an image database containing, in addition to clean image sets, the image sets corrupted by one type of noise and two-factor noises. These image sets have 69,000 images in total. Compared with existing IQA datasets, our data sets, based on CIFAR-10 and MNIST, are broadly used for deep learning networks since low-quality images reduce accuracy of neural networks. To the best of our knowledge, it is the first time that the robustness of IQAs on image sets with so many (69) perturbation conditions have been measured.

2) The research demonstrates that the BRISQUE algorithm underperforms PSNR when testing low resolution images corrupted by single or two-factor corruptions. To the best of our knowledge, such results have not previously been reported.

3) We discovered that the tested IQA algorithms do poorly at evaluating image quality of rotated images.

The rest of this paper is organized as follows. Section II introduces previously reported related results. Section III presents our design and methodology. Section IV provides our results. Further discussion and conclusions are in Sections V and VI, respectively.

XXX-X-XXXX-XXXX-X/XX/$XX.00 ©20XX IEEE

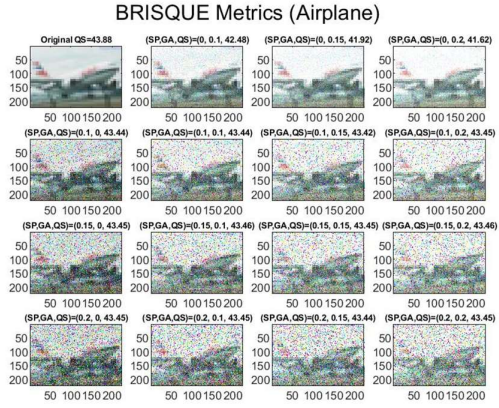

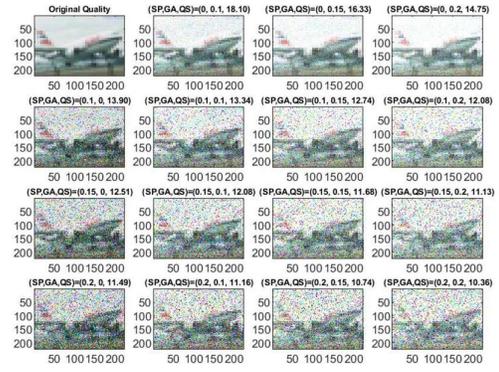

Figure 2. Results of the Gaussian-SP corruption and BRISQUE scores. The original picture is from CIFAR-10, and its file id is 376. Original QS shows the original quality score. (GA, SP, QS) stands for Gaussian noise levels, Salt & Pepper noise levels, and image quality scores of BRISQUE algorithm, respectively. X-axis and Y-axis are pixel coordinates.

Figure 3. PSNR scores of CIFAR-10_376 after Gaussian-SP corruption. The original picture is from the CIFAR-10, and its ID is 376. The PSNR is a full-reference algorithm, and the original image has zero noise and hence its score is undefined. (GA, SP, QS) denotes Gaussian noise levels, Salt & Pepper noise levels, and quality scores, respectively, X-axis and Y-axis are pixel coordinates

## II. RELATED WORK

### A. Image Quality Assessment Algorithms

Mainstream IQA algorithms include full-reference IQA algorithms and no-reference IQA algorithms. For a full-reference IQA we measure the quality of the input image by comparing it with an original reference image, while for a no-reference IQA we measure the quality of the input image without any reference image.

In this study, we focus on the Peak Signal to Noise Ratio ("PSNR" for short) [13] and the Blind/Referenceless Image Spatial Quality Evaluator ("BRISQUE" for short) [14]. PSNR is a full-reference IQA method, but BRISQUE is a no-reference IQA method.

### B. Image Quality Assessment Databases

Existing research projects evaluated IQA algorithms based on 40 IQA databases [8]. These IQA data sets consist of high-resolution images with limited numbers of pristine images. For example, both the TID2008 and TID2013 data sets have 50 pristine images with 512×384 pixels per image. In addition, the biggest IQA database, the Waterloo Exploration Database, includes 4,744 pristine images [15]. In contrast, the CIFAR-10 image set includes 60,000 32×32 pixels color pristine images, and the MNIST image set consists of 70,000 28×28 pixels grayscale pristine images. Thus compared with these deep learning image sets, the conventional IQA databases are small.

### C. One/Two-Fold Artifacts

Singly and multiply distorted IQA datasets can be used for measuring IQA algorithms [8]. For example, the MDID data set includes five corruptions: JPEG compression, JPEG2000 compression, white noise (WN) and Gaussian noise [16]. However, these existing multiply distorted databases have not included salt and pepper noise (SP for short) and rotation (RO) corruption conditions as we do here.

## III. RESEARCH DESIGN AND METHODOLOGY

In previous work [17], we assessed and compared the VGG, ResNet and AlexNet algorithms using the mCV (**m**ean accuracy and **C**oefficient of **V**ariation) graph method. In the present article, we address the BRISQUE and PSNR algorithms using box plots to assess, compare, and identify trends.

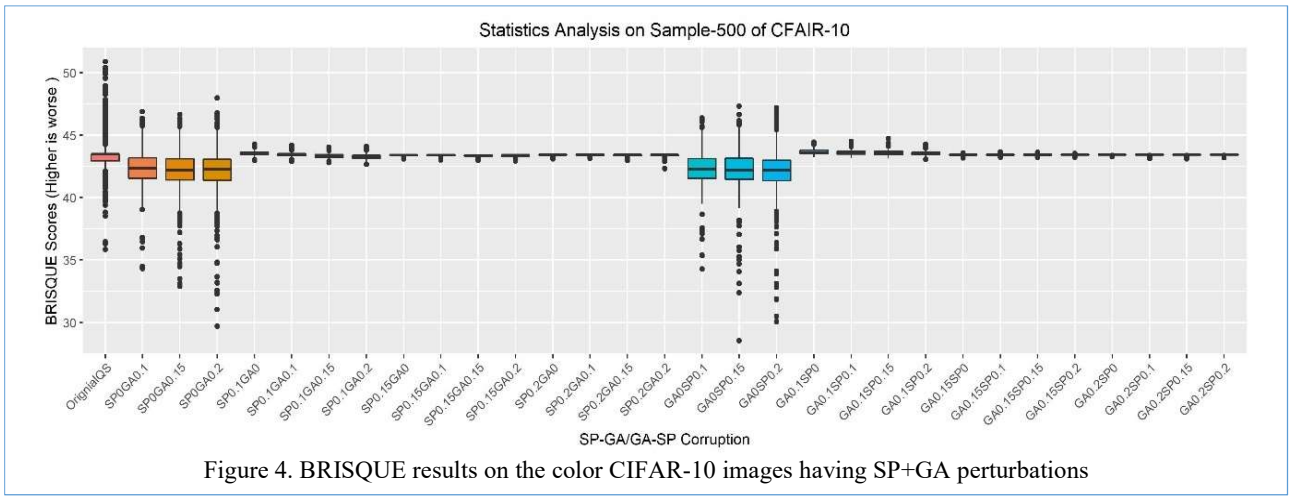

Figure 4. BRISQUE results on the color CIFAR-10 images having SP+GA perturbations

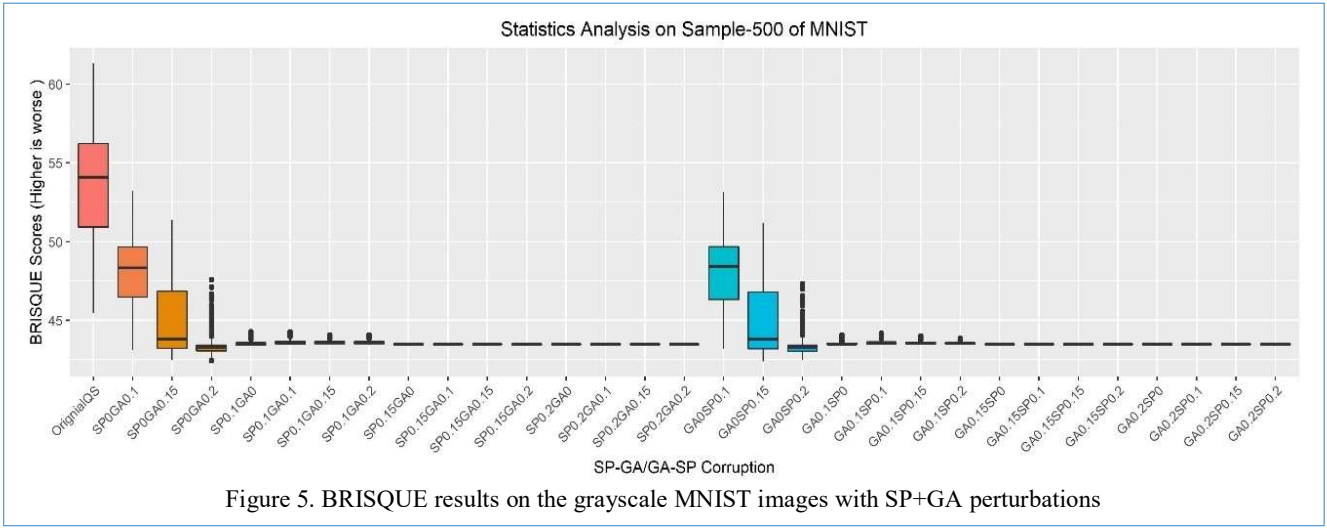
Figure 5. BRISQUE results on the grayscale MNIST images with SP+GA perturbations

## A. Two-Factor Corrupted Images

Two-factor corrupted images are images to which are applied two types of perturbation, one after the other. For example, an image is perturbed with salt & pepper noise first, then the noised image is injected additional noisy by the Gaussian algorithm. Another example is that the Gaussian noise strength could be increased from 0.1 to 0.2, then the rotation angle changed from $-60^0$ (counterclockwise) to $60^0$. Negative degrees means rotation counterclockwise, and positive degrees indicate clockwise. For details, see Tables 1 and 2.

TABLE 1. NOISE SEQUENCES, TYPES AND INDEXES

| Noise Sequences | Respective Noise Indexes | |
|---|---|---|
| SP-GA | 0.1, 0.15 & 0.2 | 0.1, 0.15 & 0.2 |
| GA-SP | 0.1, 0.15 & 0.2 | 0.1, 0.15 & 0.2 |
| SP-Rotation | 0.1, 0.15 & 0.2 | $-60^0, -30^0, 0^0, 30^0$ & $60^0$ |
| Rotation-SP | $-60^0, -30^0, 0^0, 30^0$ & $60^0$ | 0.1, 0.15 & 0.2 |

Note that SP stands for salt & pepper noise. GA stands for Gaussian noise.

## B. Data Preparation

In the study, we randomly chose 500 images from CIFAR-10 and MNIST, respectively. This gave a total of 1000 pristine images. Each image was then corrupted by 68 perturbation conditions, resulting in the original image sets, single factor noised image sets, and two-factor noised image sets. For details, see Table 2. Therefore, a total of 69,000 images were measured by the BRISQUE and PSNR algorithms.

## IV. EXPERIMENTAL RESULTS

BRISQUE and PSNR were chosen to measure image quality quantitatively. BRISQUE is a no-reference index, but PSNR is a reference index.

From Figures 4-7 and Tables 3-6, we can see:

1. The BRISQUE algorithm does not work well on single factor SP or GA corrupted images (Figs. 4 and 5). A higher BRISQUE score indicates poorer image quality, however experimental results sometimes showed an opposite trend. For instance, BRISQUE scores could be reduced when increasing the level of SP or GA noise.

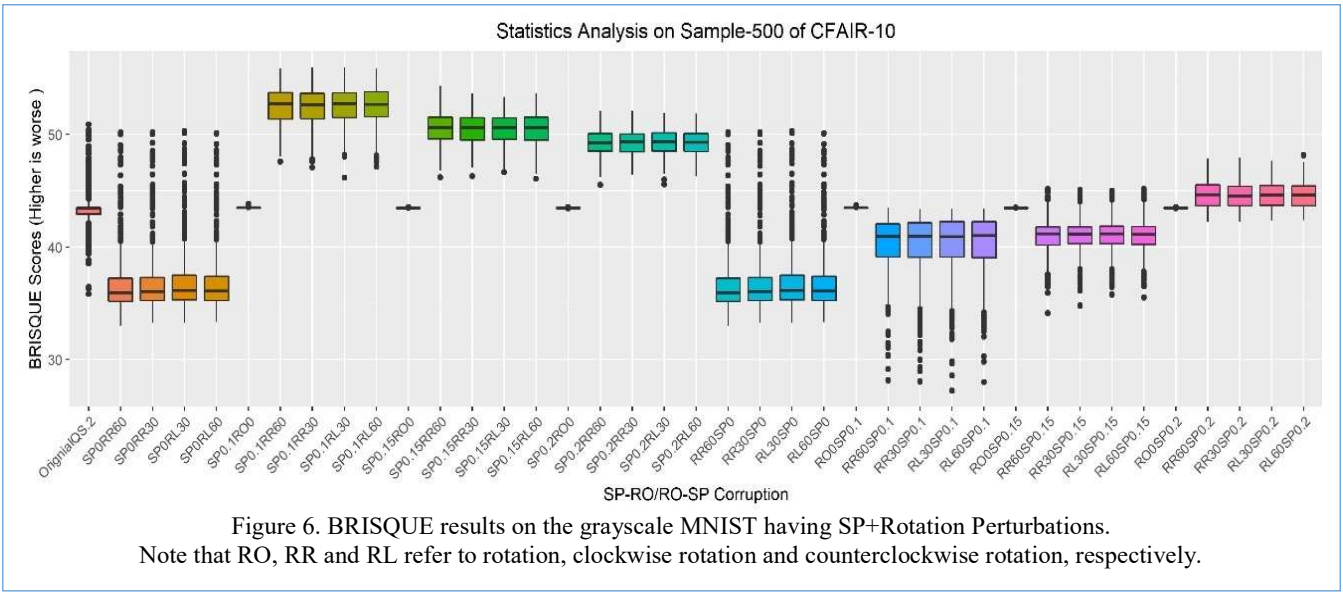
Figure 6. BRISQUE results on the grayscale MNIST having SP+Rotation Perturbations.
Note that RO, RR and RL refer to rotation, clockwise rotation and counterclockwise rotation, respectively.

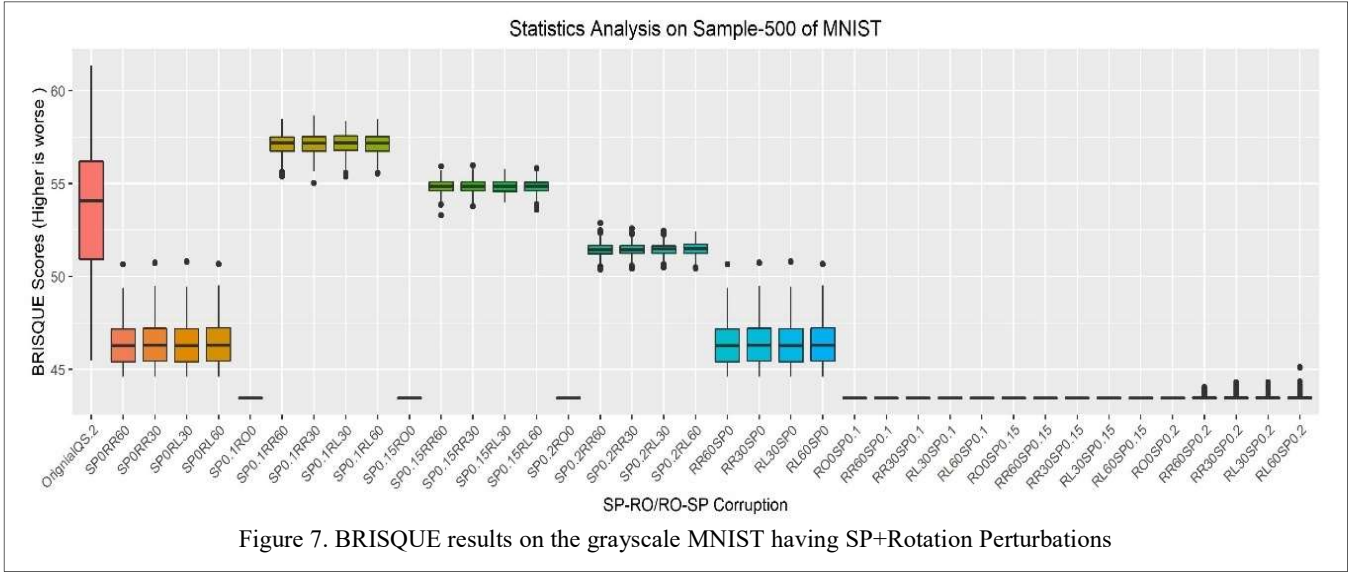

Figure 7. BRISQUE results on the grayscale MNIST having SP+Rotation Perturbations

2. BRISQUE does not perform well on two-factor SP+GA corrupted images (Figs. 4 and 5). The average BRISQUE scores of the SPxGAy corruptions and the GAxSPy corruptions on the 500-sample CIFAR-10 set are equal to 43.

3. BRISQUE does not perform well when measuring rotated images (Figs. 6 and 7). When images are rotated from zero to different angles, the average BRISQUE scores are reduced, which means that BRISQUE believes the quality of the images improved. This defies intuition.

4. BRISQUE shows limitations when measuring SP+RO (salt and paper, then rotation) corruptions (Figs. 6 and 7). When measuring left rotated images, the average scores of BRISQUE are higher than these scores of right rotated images. For instance, the BRISQUE scores of the SP0.2ROx is 48.09, which is higher than the scores of the corresponding ROxSP0.1, 44.37.

From Figures 8-11 and Tables 7-10, we can see:

1. PSNR performs as expected on single factor or two-factor SP+GA corruptions on both color CIFAR-10 sets and monochrome MNIST sets (Figs. 8 and 10).

2. PSNR does not work well on rotation (RO) corruptions or two-factor SP+RO corruptions. The average PSNR scores on two-factor SP+RO corruptions are close to each other. For example, average PSNR scores are 8.07, 9.37, 8.97, and 8.67 when measuring SP0ROx, SP0.1ROx, SP0.15ROx and SP0.2ROx corruptions, respectively.

V. DISCUSSION

In this research project, experimental performance of IQA algorithms on corrupted image sets proved that BRISQUE and PSNR have limitations on singly or multiply perturbations. We found that

1) PSNR does not correctly measure the quality of rotated image sets even though it works well on SP and/or GA corrupted images (Figs. 9 and 11). PSNR is a no-reference algorithm because it compares the pixels between original image and the corrupted images at the same locations. Rotation changes pixel positions, confusing the algorithm.

2) Our results (Figs. 8 and 10) showed that, on images noised by SPxGAy corruption conditions, more corruption, whether SP or GA, worsens the assessed image quality as assessed by PSNR.

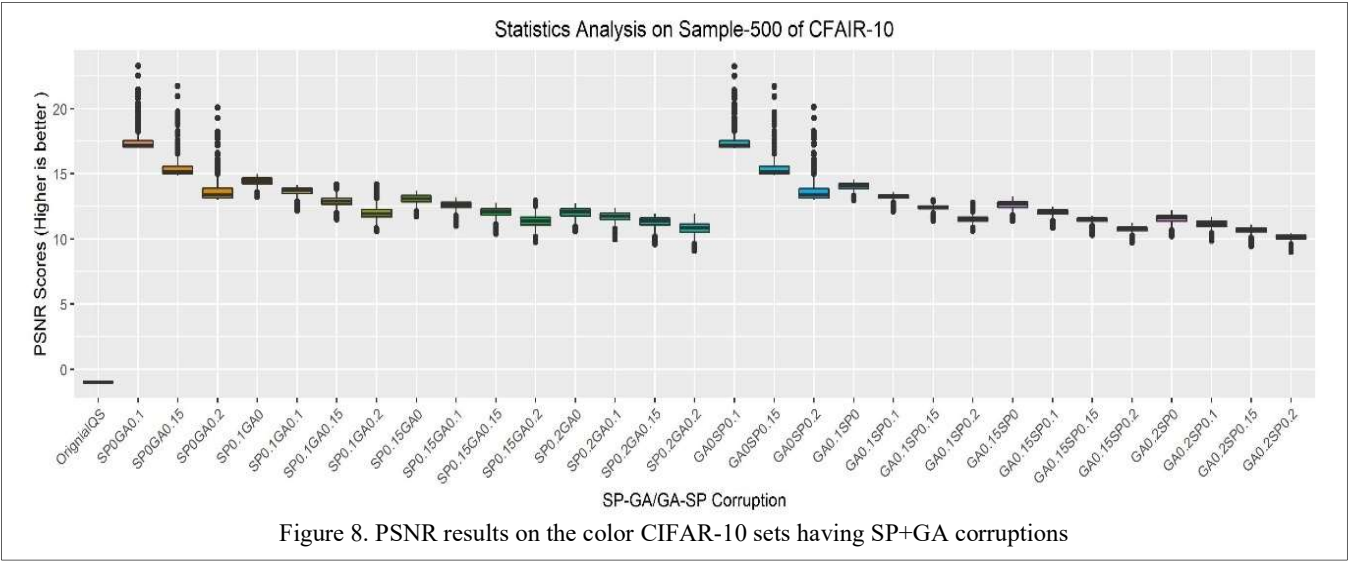

Figure 8. PSNR results on the color CIFAR-10 sets having SP+GA corruptions

3) The full reference algorithm PSNR is better than BRISQUE when measuring single factor or two-factor SP+GA corrupted image sets. Also, BRISQUE does not correctly measure the quality of rotated image sets. Our results indicate that the BRISQUE algorithm has room for performance improvement. The experimental results indicate that IQA algorithms do not perform well under two-factor SP and GA or rotation conditions. There are three possible reasons why BRISQUE does not perform well. (1) Image resolution: BRISQUE was trained and evaluated on higher resolution image sets, such as 512×768 pixels. However, the image sets of CIFAR-10 and MNIST have only 32×32 pixels. (2) Training data sets: BRISQUE was trained on TID2008. The training data consists of high-resolution, high-quality image sets without rotated images. (3) Two factor corruptions (for example) may cause the amount of error applied to each pixel to be distributed in a non-Gaussian manner. However the BRISQUE algorithm assumes a generalized Gaussian distribution on images. While BRISQUE detects a broader range of image distortion statistics than just Gaussian, nevertheless such unbalanced distributions reduce the accuracy of BRISQUE.

## VI. Conclusions

This study serves as a proof of concept for quantitatively benchmarking image quality assessment (IQA) algorithms on low resolution, one/two-fold noised image sets. In this paper, we provided a novel approach to measuring the robustness of IQA algorithms: Two-factor corrupted image sets, color CIFAR-10 and monochrome MNIST.

We evaluated two IQA algorithms on image sets that were corrupted by two-factor perturbation. The data sets had 69 benchmarking sets: an original image set, sets with single factor noise corruptions, and sets with two factor noise corruptions.

The research discovered that existing IQA algorithms have space to improve stability. BRISQUE is less good than PSNR when measuring two-factor corrupted image sets. We discussed three possible reasons about why BRISQUE does not perform well. In this experimental investigation, we discovered that both BRISQUE and PSNR do not perform well when benchmarking rotated image sets. These results suggest goals for the design of the next generation of IQA algorithms.

Acknowledgment

The study was supported in part by Google.

References

[1] K. F. Kee, "Three critical matters in big data projects for e-science: different user groups, the mutually constitutive perspective, and virtual organizational capacity," in *2015 IEEE International Conference on Big Data* (*BigData 2015*), 2015, pp. 2091–2097.

[2] J. S. Saltz and I. Shamshurin, "Achieving Agile big data science: the evolution of a team's Agile process methodology," in *2019 IEEE International Conference on Big Data* (*BigData 2019*), p. 3477-3485.

[3] S. Kaisler, F. Armour, J. A. Espinosa, and W. Money, "Big data: Issues and challenges moving forward," in *2013 46th Hawaii International Conference on System Sciences*, 2013, p. 995-1004.

[4] D. A. Asamoah, R. Sharda, A. Hassan Zadeh, and P. Kalgotra, "Preparing a data scientist: a pedagogic experience in designing a big data analytics course," *Decision Sciences Journal of Innovative Education*, vol. 15, no. 2, pp. 161-190, 2017.

[5] S. J. Russell, *Artificial Intelligence a Modern Approach*. Pearson Education, Inc., 2010.

[6] W. Dai, "Benchmarking Deep Learning Robustness on Images with Two-Factor Corruption." ProQuest Dissertations Publishing, 2020.

[7] S. Dodge and L. Karam, "Understanding how image quality affects deep neural networks," in *2016 Eighth International Conference on Quality of Multimedia Experience (QoMEX)*, 2016, pp. 1-6.

[8] G. Zhai and X. Min, "Perceptual image quality assessment: a survey," *Science China Information Sciences*, vol. 63, no. 11. 2020. doi: 10.1007/s11432-019-2757-1.

[9] N. Ponomarenko, L. Jin, O. Ieremeiev, V. Lukin, *et al.*, "Image database TID2013: peculiarities, results and perspectives," *Signal Process. Image Commun.*, vol. 30, 2015, doi: 10.1016/j.image.2014.10.009.

[10] N. Ponomarenko, V. Lukin, A. Zelensky, K. Egiazarian, *et al.*, "TID2008 - a database for evaluation of full-reference visual quality assessment metrics," *Advances of Modern Radioelectronics*, vol. 10, p. 30-45, 2009.

[11] D. Jayaraman, A. Mittal, A. K. Moorthy, and A. C. Bovik, "Objective quality assessment of multiply distorted images," 2012 *Conference Record of the Forty Sixth Asilomar Conference on Signals, Systems and Computers* (*ASILOMAR*), p. 1693-1697 doi: 10.1109/ACSSC.2012.6489321.

[12] K. Gu, G. Zhai, X. Yang, and W. Zhang, "Hybrid no-reference quality metric for singly and multiply distorted images," *IEEE Transactions on Broadcasting*, vol. 60, no. 3, 2014, doi: 10.1109/TBC.2014.2344471.

[13] A. Horé and D. Ziou, "Image quality metrics: PSNR vs. SSIM," *2010 20th International Conference on Pattern Recognition*, p. 2366-2369, doi: 10.1109/ICPR.2010.579.

[14] A. Mittal, A. K. Moorthy, and A. C. Bovik, "No-reference image quality assessment in the spatial domain," *IEEE Transactions on Image Processing*, vol. 21, no. 12, pp. 4695-4708, 2012.

[15] K. Ma, Z. Duanmu, Q. Wu, Z Wang, *et al.*, "Waterloo exploration database: new challenges for image quality assessment models," *IEEE Transactions on Image Processing*, vol. 26, no. 2, 2017, p. 1004-1016, doi: 10.1109/TIP.2016.2631888.

[16] W. Sun, F. Zhou, and Q. Liao, "MDID: A multiply distorted image database for image quality assessment," *Pattern Recognition*, vol. 61, 2017, doi: 10.1016/j.patcog.2016.07.033.

[17] W. Dai and D. Berleant, "Benchmarking robustness of deep learning classifiers using two-factor perturbation," in *2021 IEEE International Conference on Big Data* (*BigData 2021*), doi: 10.1109/BigData52589.2021.9671976.

TABLE 2. CORRUPTION TYPES.

| C1 | C2 | C3 | C4 | C5 | C6 | C7 | C8 | C9 |
|---|---|---|---|---|---|---|---|---|
| clean | SP0GA0.1 | SP0GA0.15 | SP0GA0.2 | SP0.1GA0 | SP0.1GA0.1 | SP0.1GA0.15 | SP0.1GA0.2 | SP0.15GA0 |
| C10 | C11 | C12 | C13 | C14 | C15 | C16 | C17 | C18 |
| SP0.15GA0.1 | SP0.15GA0.15 | SP0.15GA0.2 | SP0.2GA0 | SP0.2GA0.1 | SP0.2GA0.15 | SP0.2GA0.2 | GA0SP0.1 | GA0SP0.15 |
| C19 | C20 | C21 | C22 | C23 | C24 | C25 | C26 | C27 |
| GA0SP0.2 | GA0.15SP0 | GA0.15SP0.1 | GA0.15SP0.15 | GA0.15SP0.2 | GA0.1SP0 | GA0.1SP0.1 | GA0.1SP0.15 | GA0.1SP0.2 |
| C28 | C29 | C30 | C31 | C32 | C33 | C34 | C35 | C36 |
| GA0.2SP0 | GA0.2SP0.1 | GA0.2SP0.15 | GA0.2SP0.2 | SP0RR30 | SP0RR60 | SP0.1RR30 | SP0.1RR60 | SP0.15RR30 |
| C37 | C38 | C39 | C40 | C41 | C42 | C43 | C44 | C45 |
| SP0.15RR60 | SP0.2RR30 | SP0.2RR60 | SP0RL30 | SP0RL60 | SP0.1RO0 | SP0.1RL30 | SP0.1RL60 | SP0.15RO0 |
| C46 | C47 | C48 | C49 | C50 | C51 | C52 | C53 | C54 |
| SP0.15RL30 | SP0.15RL60 | SP0.2RO0 | SP0.2RL30 | SP0.2RL60 | RR30SP0.1 | RR30SP0.15 | RR30SP0.2 | RR30SP0 |
| C55 | C56 | C57 | C58 | C59 | C60 | C61 | C62 | C63 |
| RR60SP0.1 | RR60SP0.15 | RR60SP0.2 | RR60SP0 | RO0SP0.1 | RO0SP0.15 | RO0SP0.2 | RL30SP0 | RL30SP0.1 |
| C64 | C65 | C66 | C67 | C68 | C69 | | | |
| RL30SP0.15 | RL30SP0.2 | RL60SP0 | RL60SP0.1 | RL60SP0.15 | RL60SP0.2 | | | |

NOTE THAT C1 REPRESENTS A CLEAN IMAGE GROUP OR ORIGINAL IMAGE GROUP, WHEREAS C5 DENOTES AN IMAGE GROUP WHOSE IMAGES WERE FIRST SP NOISE CORRUPTED, THEN GA NOISE CORRUPTED.

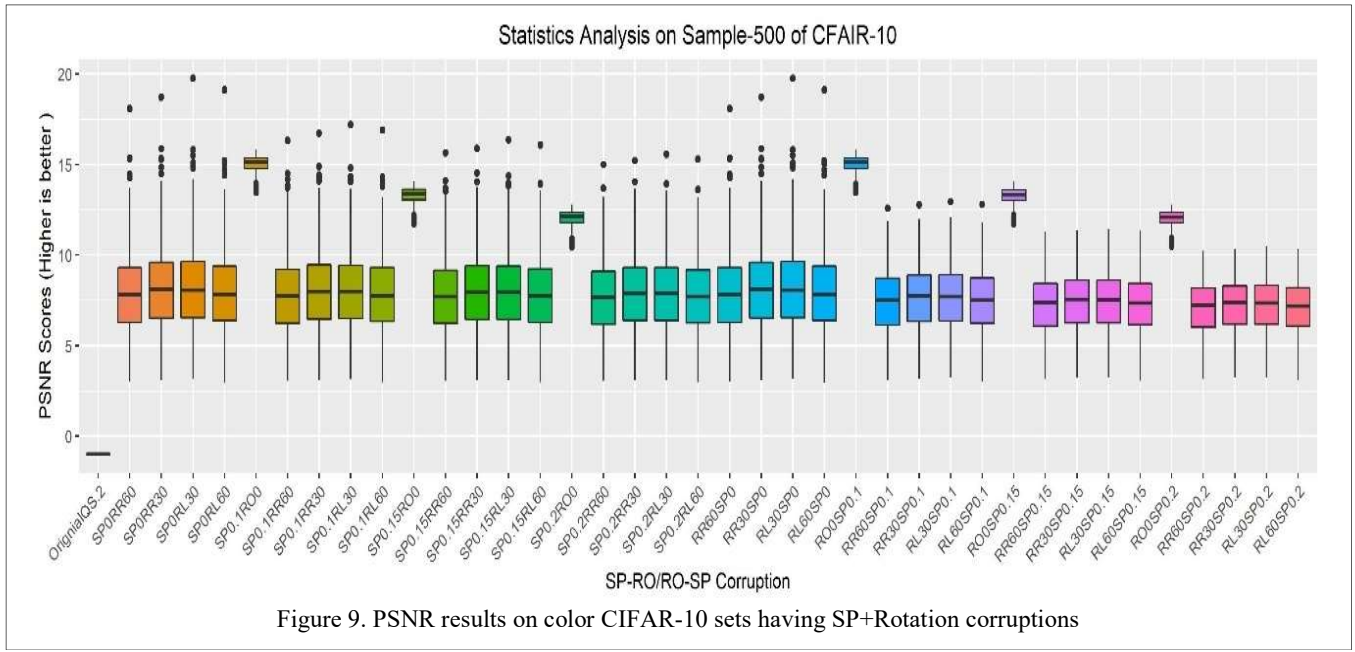

Figure 9. PSNR results on color CIFAR-10 sets having SP+Rotation corruptions

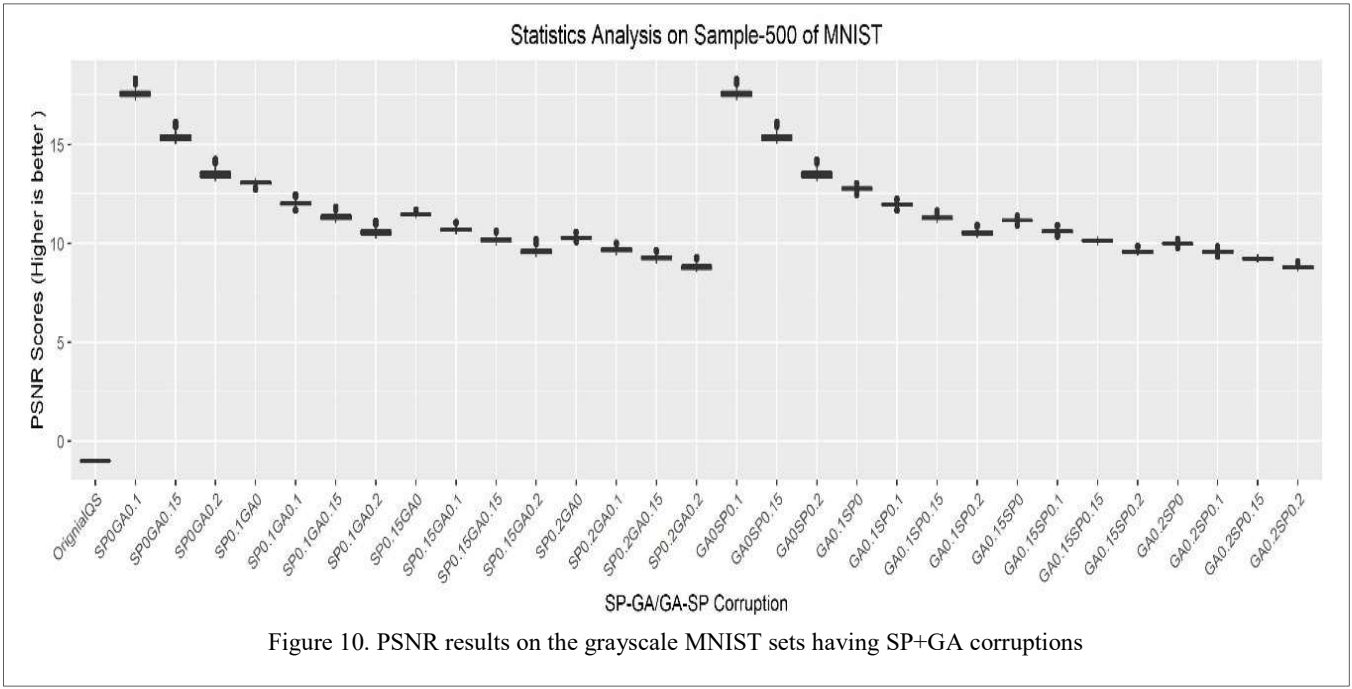

Figure 10. PSNR results on the grayscale MNIST sets having SP+GA corruptions

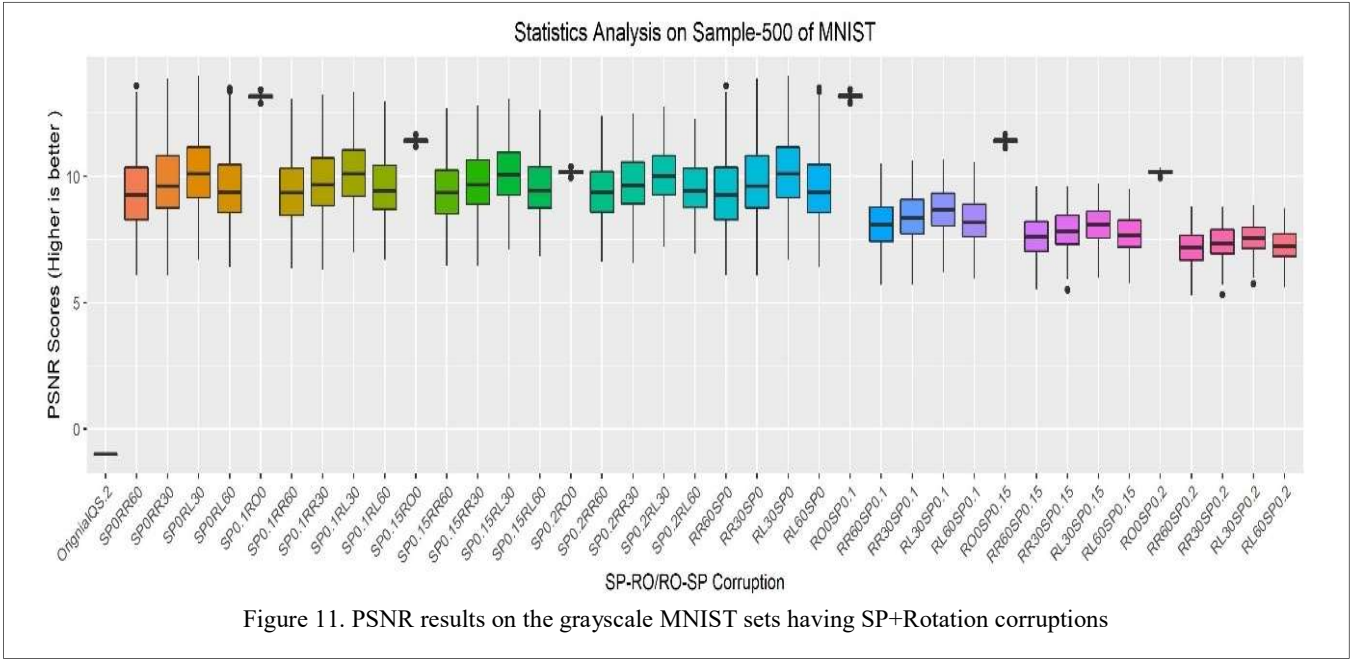

Figure 11. PSNR results on the grayscale MNIST sets having SP+Rotation corruptions

TABLE 3. AVERAGE BRISQUE SCORES WITH SP+GA CORRUPTIONS ON 500-SAMPLE SET FROM CIFAR-10.

|        | Original-QS | Avg-SP0GAx | Avg-SP0.1GAx | Avg-SP0.15GAx | Avg-SP0.2GAx |
|--------|-------------|------------|--------------|---------------|--------------|
| Min    | 35.84       | 32.30      | 42.84        | 43.01         | 42.89        |
| 1st Qu | 42.93       | 41.45      | 43.28        | 43.33         | 43.38        |
| Median | 43.46       | 42.27      | 43.38        | 43.38         | 43.42        |
| Mean   | 43.48       | 42.19      | 43.39        | 43.36         | 43.40        |
| 3rd Qu | 43.46       | 43.12      | 43.50        | 43.42         | 43.44        |
| Max    | 50.89       | 47.16      | 44.15        | 43.48         | 43.47        |
|        | Avg-GAxSP0  | Avg-GAxSP0.1 | Avg-GAx SP0.15 | Avg-GAx SP0.2 |             |
| Min    | 30.96       | 43.15      | 43.21        | 43.18         |              |
| 1st Qu | 41.45       | 43.46      | 43.41        | 43.42         |              |
| Median | 42.23       | 43.55      | 43.43        | 43.44         |              |
| Mean   | 42.17       | 43.60      | 43.42        | 43.43         |              |
| 3rd Qu | 43.08       | 43.71      | 43.45        | 43.45         |              |
| Max    | 46.95       | 44.48      | 43.58        | 43.47         |              |

TABLE 4. AVERAGE BRISQUE SCORES WITH SP+GA CORRUPTION ON THE 500-SAMPLE SET FROM MNIST.

|        | Original-QS | Avg-SP0GAx | Avg-SP0.1GAx | Avg-SP0.15GAx | Avg-SP0.2GAx |
|--------|-------------|------------|--------------|---------------|--------------|
| Min    | 45.47       | 42.68      | 43.44        | 43.46         | 43.46        |
| 1st Qu | 50.92       | 44.25      | 43.49        | 43.46         | 43.46        |
| Median | 54.09       | 45.14      | 43.55        | 43.46         | 43.46        |
| Mean   | 53.35       | 45.51      | 43.58        | 43.46         | 43.46        |
| 3rd Qu | 56.21       | 46.64      | 43.64        | 43.46         | 43.46        |
| Max    | 61.36       | 50.72      | 44.15        | 43.46         | 43.46        |
|        | Avg-GAxSP0  | Avg-GAxSP0.1 | Avg-GAx SP0.15 | Avg-GAx SP0.2 |             |
| Min    | 42.68       | 43.45      | 43.46        | 43.46         |              |
| 1st Qu | 44.18       | 43.48      | 43.46        | 43.46         |              |
| Median | 45.16       | 43.52      | 43.46        | 43.46         |              |
| Mean   | 45.50       | 43.55      | 43.46        | 43.46         |              |
| 3rd Qu | 46.64       | 43.59      | 43.46        | 43.46         |              |
| Max    | 50.55       | 44.01      | 43.46        | 43.46         |              |

TABLE 5. AVERAGE BRISQUE SCORES WITH SP+ROTATION CORRUPTIONS ON THE 500-SAMPLE SET FROM CIFAR-10.

|        | Original-QS | Avg-SP0ROx | Avg-SP0.1ROx | Avg-SP0.15ROx | Avg-SP0.2ROx |
|--------|-------------|------------|--------------|---------------|--------------|
| Min    | 35.84       | 33.21      | 46.27        | 45.73         | 45.43        |
| 1st Qu | 42.93       | 35.24      | 49.85        | 48.32         | 47.49        |
| Median | 43.46       | 36.06      | 50.84        | 49.15         | 48.13        |
| Mean   | 43.48       | 36.82      | 50.68        | 49.06         | 48.09        |
| 3rd Qu | 43.46       | 37.35      | 51.66        | 49.89         | 48.75        |
| Max    | 50.89       | 50.17      | 53.48        | 51.69         | 50.30        |
|        | Avg-ROxSP0  | Avg-ROxSP0.1 | Avg-ROx SP0.15 | Avg-ROx SP0.2 |             |
| Min    | 33.21       | 30.99      | 36.74        | 42.51         |              |
| 1st Qu | 35.24       | 39.96      | 40.89        | 43.63         |              |
| Median | 36.06       | 41.45      | 41.61        | 44.36         |              |
| Mean   | 36.82       | 40.88      | 41.49        | 44.37         |              |
| 3rd Qu | 37.35       | 42.43      | 42.13        | 45.04         |              |
| Max    | 50.17       | 43.46      | 44.76        | 47.02         |              |

TABLE 6. Average BRISQUE scores with SP+Rotation corruptions on the 500-sample set from MNIST.

|  | Original-QS | Avg-SP0ROx | Avg-SP0.1ROx | Avg-SP0.15ROx | Avg-SP0.2ROx |
|---|---|---|---|---|---|
| Min | 45.47 | 44.61 | 52.96 | 51.63 | 49.04 |
| 1st Qu | 50.92 | 45.42 | 54.10 | 52.38 | 49.69 |
| Median | 54.09 | 46.28 | 54.44 | 52.57 | 49.86 |
| Mean | 53.35 | 46.34 | 54.39 | 52.57 | 49.86 |
| 3rd Qu | 56.21 | 47.20 | 54.71 | 52.76 | 50.04 |
| Max | 61.36 | 50.73 | 55.50 | 53.40 | 50.77 |
|  | Avg-ROxSP0 | Avg-ROxSP0.1 | Avg-ROx SP0.15 | Avg-ROx SP0.2 |  |
| Min | 44.61 | 43.46 | 43.46 | 43.46 |  |
| 1st Qu | 45.42 | 43.46 | 43.46 | 43.46 |  |
| Median | 46.28 | 43.46 | 43.46 | 43.47 |  |
| Mean | 46.34 | 43.46 | 43.46 | 43.49 |  |
| 3rd Qu | 47.20 | 43.46 | 43.46 | 43.49 |  |
| Max | 50.73 | 43.46 | 43.46 | 44.24 |  |

TABLE 7. AVERAGE PSNR SCORES WITH SP+GA CORRUPTIONS ON THE 500-SAMPLE SET FROM CIFAR-10.

|  | Avg-SP0GAx | Avg-SP0.1GAx | Avg-SP0.15GAx | Avg-SP0.2GAx |
|---|---|---|---|---|
| Min | 14.94 | 11.88 | 10.72 | 9.81 |
| 1st Qu | 15.05 | 13.01 | 12.03 | 11.21 |
| Median | 15.23 | 13.26 | 12.29 | 11.51 |
| Mean | 15.53 | 13.25 | 12.25 | 11.46 |
| 3rd Qu | 15.66 | 13.48 | 12.52 | 11.75 |
| Max | 21.69 | 14.37 | 13.15 | 12.27 |
|  | Avg-GAxSP0 | Avg-GAxSP0.1 | Avg-GAx SP0.15 | Avg-GAx SP0.2 |
| Min | 14.94 | 11.79 | 10.57 | 9.64 |
| 1st Qu | 15.06 | 12.65 | 11.59 | 10.72 |
| Median | 15.23 | 12.81 | 11.75 | 10.90 |
| Mean | 15.53 | 12.79 | 11.72 | 10.86 |
| 3rd Qu | 15.66 | 12.94 | 11.89 | 11.06 |
| Max | 21.69 | 13.46 | 12.19 | 11.36 |

TABLE 8. AVERAGE PSNR SCORES WITH SP+ROTATION CORRUPTIONS ON THE 500-SAMPLE SET FROM CIFAR-10.

|  | Avg-SP0ROx | Avg-SP0.1ROx | Avg-SP0.15ROx | Avg-SP0.2ROx |
|---|---|---|---|---|
| Min | 3.06 | 5.15 | 4.80 | 4.53 |
| 1st Qu | 6.43 | 8.06 | 7.69 | 7.41 |
| Median | 7.95 | 9.32 | 8.94 | 8.66 |
| Mean | 8.07 | 9.37 | 8.97 | 8.67 |
| 3rd Qu | 9.49 | 10.56 | 10.16 | 9.85 |
| Max | 18.93 | 16.61 | 15.61 | 14.78 |
|  | Avg-ROxSP0 | Avg-ROxSP0.1 | Avg-ROx SP0.15 | Avg-ROx SP0.2 |
| Min | 3.06 | 5.20 | 4.88 | 4.66 |
| 1st Qu | 6.43 | 7.98 | 7.56 | 7.25 |
| Median | 7.95 | 9.11 | 8.63 | 8.25 |
| Mean | 8.07 | 9.06 | 8.54 | 8.13 |
| 3rd Qu | 9.49 | 10.13 | 9.54 | 9.08 |
| Max | 18.93 | 13.39 | 11.91 | 10.84 |

TABLE 9. AVERAGE PSNR SCORES WITH SP+GA CORRUPTIONS ON THE 500-SAMPLE SET FROM MNIST.

|        | Avg-SP0GAx | Avg-SP0.1GAx | Avg-SP0.15GAx | Avg-SP0.2GAx |
|--------|------------|--------------|---------------|--------------|
| Min    | 15.11      | 11.42        | 10.21         | 9.24         |
| 1st Qu | 15.32      | 11.65        | 10.40         | 9.43         |
| Median | 15.44      | 11.73        | 10.47         | 9.51         |
| Mean   | 15.47      | 11.74        | 10.48         | 9.51         |
| 3rd Qu | 15.59      | 11.82        | 10.55         | 9.58         |
| Max    | 16.20      | 12.16        | 10.87         | 9.85         |
|        | Avg-GAxSP0 | Avg-GAxSP0.1 | Avg-GAx SP0.15 | Avg-GAx SP0.2 |
| Min    | 15.12      | 11.37        | 10.14         | 9.18         |
| 1st Qu | 15.32      | 11.56        | 10.31         | 9.34         |
| Median | 15.45      | 11.63        | 10.36         | 9.39         |
| Mean   | 15.47      | 11.63        | 10.36         | 9.39         |
| 3rd Qu | 15.59      | 11.70        | 10.42         | 9.44         |
| Max    | 16.19      | 11.94        | 10.60         | 9.60         |

TABLE 10. AVERAGE PSNR SCORES WITH SP+ROTATION CORRUPTIONS ON THE 500-SAMPLE SET FROM MNIST.

|        | Avg-SP0ROx | Avg-SP0.1ROx | Avg-SP0.15ROx | Avg-SP0.2ROx |
|--------|------------|--------------|---------------|--------------|
| Min    | 6.32       | 7.85         | 7.61          | 7.46         |
| 1st Qu | 8.69       | 9.67         | 9.36          | 9.13         |
| Median | 9.59       | 10.34        | 9.99          | 9.73         |
| Mean   | 9.72       | 10.42        | 10.05         | 9.78         |
| 3rd Qu | 10.70      | 11.15        | 10.74         | 10.42        |
| Max    | 13.73      | 13.21        | 12.57         | 12.07        |
|        | Avg-ROxSP0 | Avg-ROxSP0.1 | Avg-ROx SP0.15 | Avg-ROx SP0.2 |
| Min    | 6.32       | 7.30         | 6.78          | 6.38         |
| 1st Qu | 8.69       | 8.79         | 8.10          | 7.55         |
| Median | 9.59       | 9.29         | 8.53          | 7.91         |
| Mean   | 9.72       | 9.32         | 8.54          | 7.91         |
| 3rd Qu | 10.70      | 9.87         | 9.01          | 8.30         |
| Max    | 13.73      | 11.17        | 10.02         | 9.12         |